\pdfoutput=1

\documentclass[10pt,twocolumn,letterpaper]{article}

\usepackage[pagenumbers]{cvpr} 

\definecolor{cvprblue}{rgb}{0.21,0.49,0.74}
\usepackage[pagebackref,breaklinks,colorlinks,allcolors=cvprblue]{hyperref}

\def\paperID{*****} 
\def\confName{CVPR}
\def\confYear{2025}

\title{HSEmotion Team at ABAW-8 Competition: Audiovisual Ambivalence/Hesitancy, Emotional Mimicry Intensity and Facial Expression Recognition}

\author{Andrey V. Savchenko\textsuperscript{1,2}\\
\textsuperscript{1}HSE University\\
Laboratory of Algorithms and Technologies for Network Analysis, Nizhny Novgorod, Russia\\
\textsuperscript{2}Sber AI Lab\\
Moscow, Russia\\
{\tt\small avsavchenko@hse.ru}
}

\begin{document}
\maketitle
\begin{abstract}
This article presents our results for the eighth Affective Behavior Analysis in-the-Wild (ABAW) competition. We combine facial emotional descriptors extracted by pre-trained models, namely, our EmotiEffLib library, with acoustic features and embeddings of texts recognized from speech. The frame-level features are aggregated and fed into simple classifiers, e.g., multi-layered perceptron (feed-forward neural network with one hidden layer), to predict ambivalence/hesitancy and facial expressions. In the latter case, we also use the pre-trained facial expression recognition model to select high-score video frames and prevent their processing with a domain-specific video classifier. The video-level prediction of emotional mimicry intensity is implemented by simply aggregating frame-level features and training a multi-layered perceptron.  Experimental results for three tasks from the ABAW challenge demonstrate that our approach significantly increases validation metrics compared to existing baselines.
\end{abstract}

\section{Introduction}\label{sec:intro}
Emotion recognition is crucial in advancing human-computer interaction, psychological research, and AI applications~\cite{guo2024development,khare2024emotion}. Understanding human emotionality in real-world scenarios enables the development of empathetic AI systems, enhances mental health monitoring, and improves human-centric technologies such as virtual assistants, social robots, and affective computing systems~\cite{kollias2019face,kollias2023multi}. However, analyzing affective behavior in practical applications presents unique challenges due to the complexity and variability of human emotions, which are often subtle, ambiguous, and context-dependent. For instance, in mental health, accurately identifying ambivalence or hesitancy~\cite{Kollias2025} in a patient's behavior can uncover conflicting emotions and intentions related to behavioral change interventions and provide early indicators of psychological distress or decision-making struggles. Emotional mimicry intensity (EMI), which reflects the degree to which individuals unconsciously mirror the emotions of others~\cite{Qiu_2024_CVPR,yu2024efficient}, is a critical marker of social bonding and empathy. Facial Expression Recognition (FER), on the other hand, is a foundational component of emotion AI, enabling applications such as personalized education, customer behavior analysis, and human-robot interaction~\cite{guo2024facial,pmlr-v202-savchenko23a}. However, achieving high accuracy in these tasks requires robust methods that can handle real-world data's inherent noise and variability, such as poor lighting, occlusions, and diverse cultural expressions~\cite{kollias2019deep,kollias2021affect}. This underscores the need for innovative approaches integrating multimodal data and leveraging advanced machine learning techniques to improve performance.

The series of ABAW (Affective Behavior Analysis in-the-Wild) competitions~\cite{kollias2020analysing,kollias2021analysing,kollias2022abaw,kollias2023abaw2,kollias20247th} based on Aff-Wild~\cite{zafeiriou2017aff} and Aff-Wild2~\cite{kollias2019expression} datasets has been instrumental in pushing the boundaries of emotion recognition by providing researchers with a challenging and dynamic benchmark for evaluating models in real-world, in-the-wild conditions. The eighth edition of the ABAW competition (ABAW-8)~\cite{Kollias2025,kolliasadvancements} introduces a novel task that focuses on Ambivalence/Hesitancy (AH) and continues studies of measuring EMI and video-based frame-wise Expression (EXPR) recognition from the sixth ABAW challenge~\cite{kollias20246th}, reflecting the complexity and nuances of human emotional expression. These tasks are essential for developing systems that can accurately interpret human emotions in naturalistic settings, paving the way for more empathetic and responsive technologies.

In this paper, we present the results of our team in the ABAW-8 competition. We propose a novel framework that integrates multiple modalities, visual, acoustic, and linguistic, to enhance the accuracy and robustness of affective behavior recognition. Our approach combines facial emotional descriptors extracted using the EmotiEffLib library~\cite{savchenko2023cvprw} with acoustic emotional features~\cite{baevski2020wav2vec,hsu2021hubert} and text embeddings~\cite{liu2019roberta} obtained from speech recognition~\cite{radford2023robust}. We employ simple yet effective classifiers for each feature, such as linear feed-forward neural networks, and aggregate their predictions in a late fusion technique. Additionally, we employ a pre-trained FER model to filter video frames with high confidence scores, reducing redundant processing and optimizing computational resources. Our experimental results demonstrate the effectiveness of this approach, highlighting its potential to advance the field of affective computing and emotion understanding.

\section{Related Works}
In this section, we briefly review the results of participants of previous ABAW-6 tasks~\cite{kollias20246th}, namely, EXPR classification and EMI estimation.

\subsection{EXPR Classification}
One of the most widely studied ABAW challenges is the frame-wise uni-task expression classification in video~\cite{kollias20246th,kollias2023abaw2,kollias2022abaw}. It is required to assign one of eight basic emotions to each video frame using audio-visual information and models pre-trained on external datasets. Top-performing solutions often employ pre-trained deep learning models fine-tuned on this dataset. For instance, the usage of CLIP was proposed in~\cite{lin2024robust}, where the visual features are fed into a trainable multilayer perceptron (MLP) enhanced with Conditional Value at Risk. The HSEmotion team utilized lightweight architectures like MobileViT and EfficientNet, trained in multi-task scenarios~\cite{kollias2023abaw} to recognize facial expressions, valence, and arousal on static photos~\cite{Savchenko_2024a_CVPR,Savchenko_2024b_CVPR}. These models extracted frame-level features fed into simple classifiers, such as linear feed-forward neural networks, to predict facial expressions. This approach significantly improved validation metrics compared to existing baselines. 

Yu et al.~\cite{Yu_2024_CVPR} explored semi-supervised learning (SSL) techniques to address the limited size of labeled FER datasets. By generating pseudo-labels for unlabeled faces and employing an encoder to capture temporal relationships between neighboring frames, their approach achieved third place in the ABAW-6 competition. 

The second place was achieved by Masked Autoencoders (MAE)~\cite{Zhou_2024_CVPR}, which has been pre-trained on
external datasets and then fine-tuned on the EXPR training set. It was also proposed that continuous emotion recognition be improved by integrating Temporal Convolutional Network and Transformer Encoder modules.

Finally, the winners~\cite{zhang2024effective} also employed MAE for high-quality facial feature extraction, leading to superior performance across multiple affective behavior analysis tasks. Moreover, they proposed integrating multi-modal knowledge using a transformer-based feature fusion module to combine emotional information from audio signals, visual images, and transcripts. In addition, they divided the dataset division based on scene characteristics and
trained a separate classifier for each scene.

\subsection{EMI Estimation}
EMI estimation task is a multi-label video classification problem, in which it is required to predict the emotional intensity of the video by selecting from a range of six predefined emotional categories~\cite{kollias20246th}. Successful approaches integrated multimodal features, including visual, auditory, and textual cues, to capture a comprehensive emotional profile. For example, Yu et al.~\cite{yu2024efficient} presented a third-place solution focusing on efficient feature extraction and a late fusion strategy~\cite{savchenko2012adaptive} for audiovisual EMI estimation. They extracted dual-channel visual features and single-channel audio features, averaging the predictions of visual and acoustic models to achieve a more accurate estimation. 

The second pace took a technique with pure audio analysis~\cite{Hallmen_2024_CVPR}. A pre-trained Wav2Vec2.0~\cite{baevski2020wav2vec} was used with a Valence-Arousal-Dominance (VAD) module. After extracting the acoustic features and the
VAD predictions compute a global feature vector and fuse the temporal features in a recurrent neural network. They also demonstrated that visual features in this challenge are not very useful.

Qiu et al.~\cite{Qiu_2024_CVPR} developed a language-guided multi-modal framework for EMI estimation, leveraging textual modality to enhance the estimation process. Their experiments demonstrated the effectiveness of this approach, achieving top performance in the competition.

These studies collectively highlight the advancements in FER and EMI estimation, emphasizing the importance of integrating multimodal data, leveraging pre-trained models, multi-task learning~\cite{kollias2021distribution,kollias2024distribution} and employing innovative fusion strategies to enhance the accuracy and robustness of affective behavior analysis systems.

\section{Proposed Approach}\label{sec:proposed}
\begin{figure}[t]
 \centering
 \includegraphics[width=0.99\linewidth]{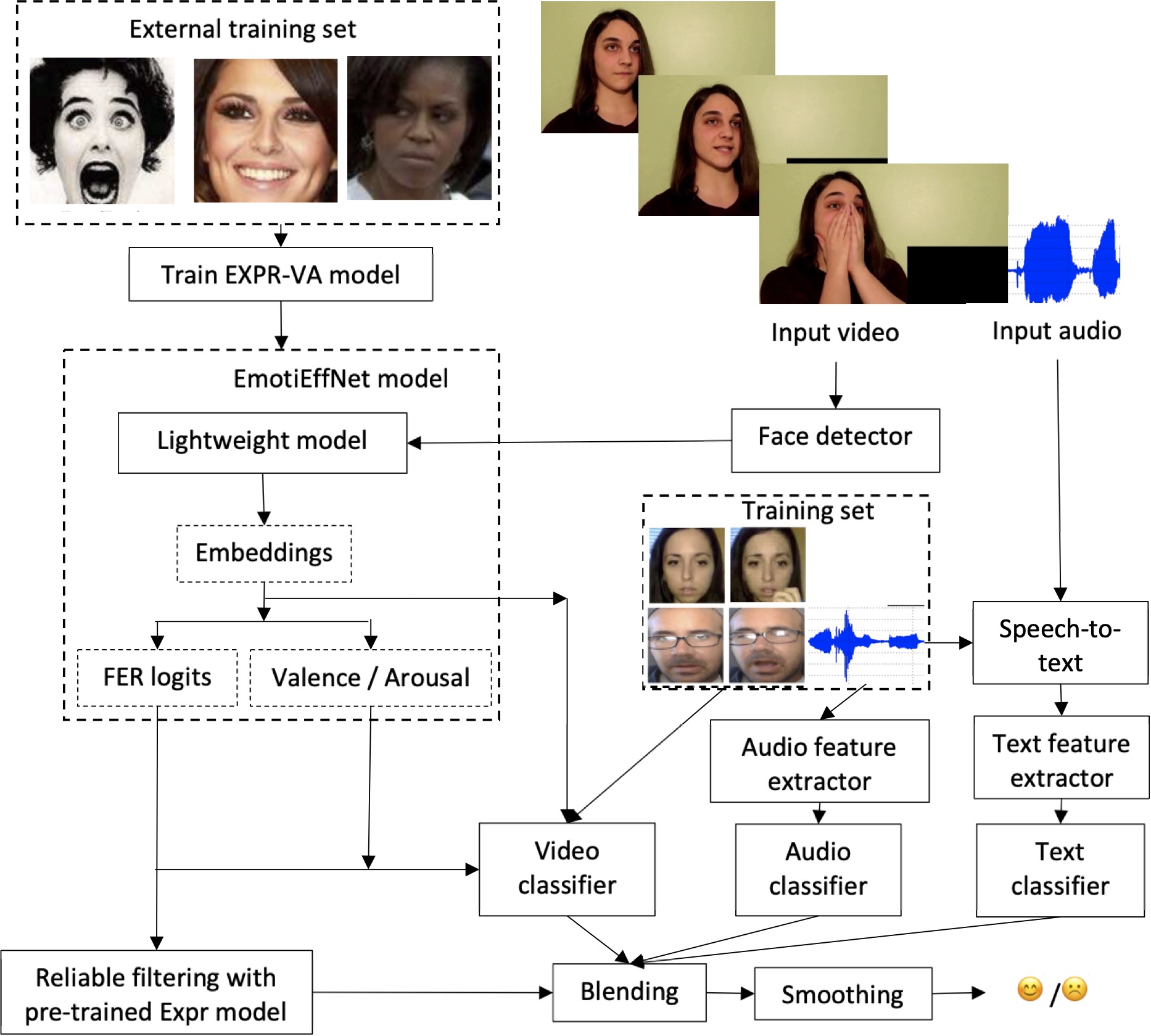}
 \caption{Proposed approach.}
 \label{fig:pipeline}
\end{figure}

In this paper, we introduce the novel approach to audio-visual affective behavior analysis (Fig.~\ref{fig:pipeline}) based on lightweight facial emotion recognition models from EmotiEffLib~\cite{Savchenko_2024b_CVPR}, a library of efficient neural networks optimized for real-time emotion recognition. Our approach is designed to leverage multimodal inputs, including visual, acoustic, and textual features, to enhance the recognition of affective behavior. The pipeline comprises several key components: feature extraction, multimodal fusion, and classification. The goal is to predict FER, EMI Estimation, and AH Recognition efficiently while maintaining computational efficiency and robustness.

At first, facial regions are cropped in each video frame using RetinaFace~\cite{deng2020retinaface} or mediapipe landmark detector~\cite{lugaresi2019mediapipe}. At the feature extraction stage, we employ our pre-trained EmotiEffLib neural networks to extract facial emotional descriptors from video frames. Our approach processes individual frames using lightweight neural network architectures~\cite{demochkina2021mobileemotiface}, such as EmotiEffNet and MobileViT, pre-trained on large-scale affective behavior datasets. These models output embeddings from the penultimate layer and scores (logits of probability distributions) over eight categories of facial expressions from AffectNet database~\cite{mollahosseini2017affectnet}, including anger, disgust, fear, happiness, sadness, surprise, neutral, and contempt.

Additionally, acoustic features are extracted using wav2vec 2.0~\cite{baevski2020wav2vec} or HuBERT~\cite{hsu2021hubert} models. Moreover, audio modality is fed into Speech-to-text module~\cite{savchenko2016information} to recognize speech using, e.g., OpenAI's Whisper small.en model~\cite{radford2023robust}. The text embeddings are generated from recognized speech using open-source models, e.g., RoBERTa~\cite{liu2019roberta} trained on GoEmotions dataset~\cite{demszky2020goemotions} or commercial ChatGPT and GigaChat embeddings. These modalities capture complementary information, ensuring a comprehensive understanding of affective states.

Each modality is classified separately using specially trained MLP. The fusion and classification step aggregates the extracted frame-wise features across time, creating video-level representations. We employ a simple yet effective linear feed-forward neural network to classify facial videos, audio, and recognized text. We perform blending of these classifiers to obtain the final predictions~\cite{savchenko2020ad}. To improve accuracy and stability, we apply temporal smoothing to the frame-wise predictions, reducing noise, preventing abrupt fluctuations in classification outputs, and capturing the natural transitions of facial expressions over time~\cite{savchenko2023cvprw}. 
Below, we provide details on the methodology for each task.

\subsection{Frame-wise Facial Expression Recognition}

Participants were tasked with recognizing eight categories of facial expressions: the seven basic emotions (anger, disgust, fear, happiness, sadness, surprise, and neutral) plus an 'other' category. The competition utilized the Aff-Wild2 dataset~\cite{kollias2019expression}, comprising 548 videos with approximately 2.7 million frames annotated for these expressions.

To extract facial embeddings, we leverage EmotiEffNet-B0 models, which have shown the best performance among EmotiEffib models~\cite{Savchenko_2024b_CVPR}. It extracts 1280-dimensional embeddings, which we fed into trainable MLP with 128 units in hidden layers. A similar MLP is trained on top of wav2vec 2.0 acoustic features. We do not use speech-to-text and text embeddings here, as most videos do not contain large samples of recognizable speech. The bending of video and audio MLPs outputs (8 class posterior probabilities) is performed using a weight hyperparameter $w \in [0,1]$.

To enhance computational efficiency and quality, we integrate the same pre-trained EmotiEffNet-B0 model that selects frames with high confidence scores, reducing unnecessary processing. We estimate the posterior probabilities of basic emotions by computing softmax to the logits at the output of the final layer. If the maximal probability is relatively high (greater than a predefined threshold $t$), we return this expression without using trained MLP for video and audio classifiers. Such a filtering mechanism selects frames with high confidence scores, preventing redundant or misleading frames from affecting the final classification.

\subsection{Emotional Mimicry Intensity Estimation}
The EMI Estimation challenge aimed to assess the intensity of participants' emotional mimicry in response to 'seed' videos displaying specific emotions (stimulus). Participants were required to predict the intensities of six self-reported emotions for the entire video: admiration, amusement, determination, empathic pain, excitement, and joy. The multimodal Hume-Vidmimic2 dataset, consisting of over 15,000 videos totaling more than 30 hours of audiovisual content, was employed for this task.

We extract multimodal features for facial expressions, voice, and speech content. For visual features, we utilize MT-EmotiMobileFaceNet or MT-EmotiMobileViT~\cite{Savchenko_2024a_CVPR} to capture fine-grained emotional expressions. We used the facial images officially provided by the organizers. The best results are obtained for the classification of logits at the output of the final layer. We compute STAT (statistical) features (component-wise mean, std, min, max)~\cite{bargal2016emotion} of frame-wise logits to obtain a 28-dimensional video descriptor. 

Audio signals are processed using wav2vec 2.0~\cite{baevski2020wav2vec} and HuBERT~\cite{hsu2021hubert}, which generate embeddings that reflect prosody, tone, and intonation. The text modality, extracted using RoBERTa, ChatGPT (small model), or GigaChat, captures the semantic content of spoken words. 

The MLPs with six sigmoid outputs are trained for each modality to optimize the weighted Pearson correlations, where the weights are inversely proportional to class counts. The outputs of MLPs are weighted using late fusion techniques, where each modality contributes to the final EMI estimation based on its reliability. 

\subsection{Ambivalence/Hesitancy Recognition}

The BAH (Behavioural Ambivalence/Hesitancy) dataset~\cite{Kollias2025} identifies ambivalence and hesitancy in Q\&A videos, which are recorded explicitly for behavioral analysis. The recorded subjects were asked seven questions to elicit a range of emotional responses, including neutral, positive, negative, ambivalent, willing, resistant, or hesitant feelings about their behaviors. The training and validation sets include 84 participants with up to 7 videos, totaling 431 videos with a combined duration of 3.4 hours and approximately 295,500 frames. Each video frame is annotated by the presence (1) or absence (0) of AH. 

Our model extracts facial, vocal, and textual features using the general multimodal approach (Fig.~\ref{fig:pipeline}). A crucial step in our pipeline is the fusion of textual embeddings with visual and acoustic cues, as verbal content often plays a key role in expressing ambivalence. We employ early fusion strategies where the embeddings from RoBERTa trained on GoEmotions~\cite{demszky2020goemotions}, wav2vec 2.0~\cite{baevski2020wav2vec}, and MT-EmotiMobileFaceNet~\cite{Savchenko_2024a_CVPR} are concatenated and processed by a feed-forward neural network. Additionally, we experiment with blending techniques, where predictions from individual modalities are averaged to refine the final classification output.

Unfortunately, the acoustic and text features are not aligned with video frames. To perform frame-level predictions, we interpolated the acoustic and text features for the shape of visual features using interp1d from SciPy. Moreover, we examined the possibility of training a video-level classifier to predict if it contains at least one frame with an AH label set to 1. Here, we compute the component-wise mean of text features, train a logistic regression model, and use its output to filter videos for which this classifier predicts an absence of AH.

\section{Experimental Results}\label{sec:exper}

\begin{table*}
 \centering
 \begin{tabular}{ccccc}
 \toprule
 Method & Modality & Is ensemble? & F1-score $P_{EXPR}$ & Accuracy \\
 \midrule
 Baseline VGGFACE (MixAugment)~\cite{kollias20246th} & Faces & No & 25.0 & -\\
\hline
EfficientNet-B0~\cite{savchenko2022cvprw} & Faces & No & 40.2 & - \\
MT-EmotiMobileFaceNet~\cite{Savchenko_2024a_CVPR} & Faces & No & 32.7 & 46.2 \\
MT-EmotiMobileViT~\cite{Savchenko_2024a_CVPR} & Faces & No & 35.6 & 46.1 \\
Meta-Classifier~\cite{wang2023facial} & Faces & Yes & 30.2 & 46.2 \\ 
CLIP~\cite{lin2024robust} & Faces & No & 36.0 & - \\ 
TCN~\cite{zhou2023continuous} & Audio/video & Yes& 37.7 & -  \\
Transformer~\cite{zhang2023facial} & Audio/video & Yes & 40.6 & - \\ 
SSL + Temporal+ Post-process~\cite{Yu_2024_CVPR}& Audio/video & Yes & 44.43 & - \\
MAE~\cite{zhang2023abaw5} & Audio/video & Yes & 49.5 & - \\
MAE+Transformer feature fusion~\cite{zhang2024effective} & Audio/video & Yes & 55.55 & - \\ 
\hline
wav2vec 2.0 & Audio & No & 29.09 & 41.01 \\
wav2vec 2.0, smoothing & Audio & No & 35.95 & 52.36 \\
EmotiEffNet& Faces & No & 38.44 & 49.54 \\
EmotiEffNet, smoothing & Faces & No & 42.37 & 54.34 \\
EmotiEffNet, filtering + smoothing & Faces & No & 43.83 &54.29 \\
wav2vec 2.0+EmotiEffNet& Audio/video & Yes & 40.30 & 52.03 \\
wav2vec 2.0+EmotiEffNet, smoothing & Audio/video & Yes & 43.43 & 55.67 \\
wav2vec 2.0+EmotiEffNet, filtering + smoothing &Audio/video & Yes &  44.59 & 55.32 \\
 \bottomrule
 \end{tabular}
 \caption{Expression Challenge Results on the Aff-Wild2’s validation set.}
 \label{tab:expr}
\end{table*}

\subsection{Facial Expression Recognition}
To evaluate our approach on the EXPR classification task, we tested various models on the Aff-Wild2 validation set. Table~\ref{tab:expr} summarizes the F1-scores and accuracy achieved by different methods. The baseline VGGFACE (MixAugment) model achieved an F1-score of 25.0, while EfficientNet-B0 improved this score to 40.2.

Our best-performing model, EmotiEffNet with filtering and smoothing, achieved an F1-score of 43.83\%, outperforming previous approaches that used visual models from EmotiEffLib~\cite{Savchenko_2024b_CVPR} up to 1.5\%. The combination of wav2vec 2.0 for audio processing~\cite{baevski2020wav2vec} and EmotiEffNet for FER~\cite{savchenko2023cvprw} further improved the performance, reaching an F1-score of 44.59\% if filtering with pre-trained EMotiEffNet-B0 model is applied.

\begin{figure}[t]
 \centering
 \includegraphics[width=0.95\linewidth]{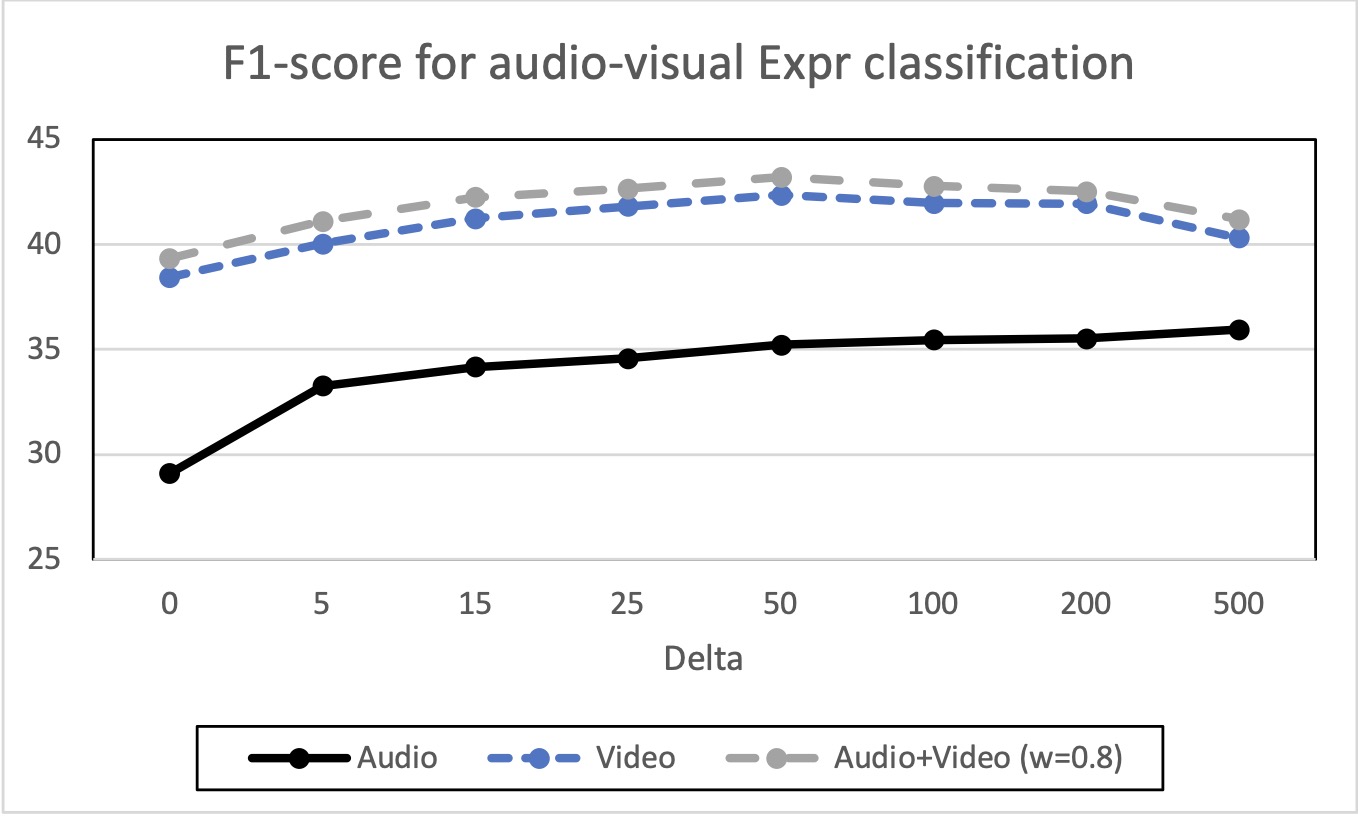}
 \caption{Dependence of F1-score for video Expr recognition on the smoothing kernel size $k$.}
 \label{fig:expr_smoothing}
\end{figure}

\begin{figure}[t]
 \centering
 \includegraphics[width=0.95\linewidth]{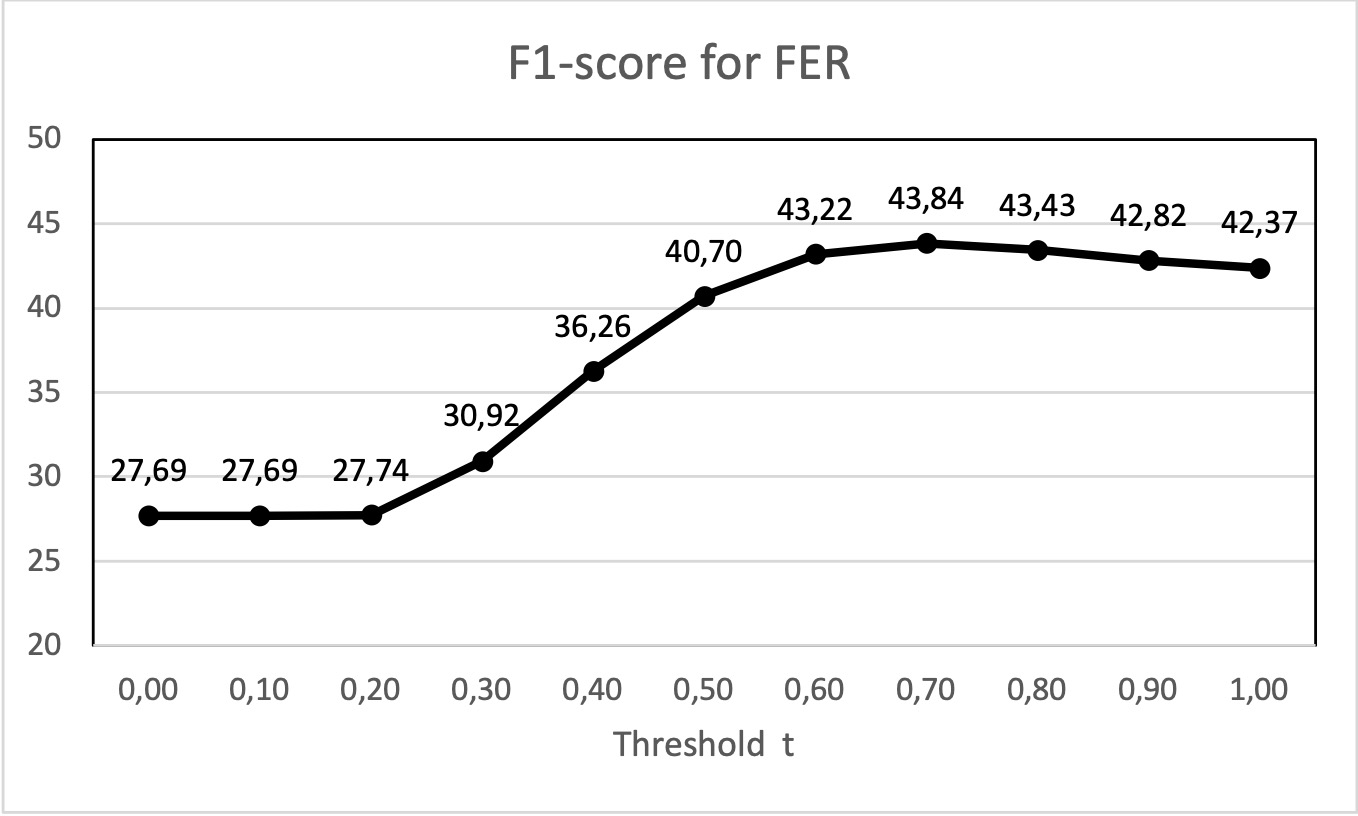}
 \caption{Dependence of F1-score for video Expr recognition on the filtering threshold $t$.}
 \label{fig:expr_video_thresholding}
\end{figure}

\begin{figure}[t]
 \centering
 \includegraphics[width=0.95\linewidth]{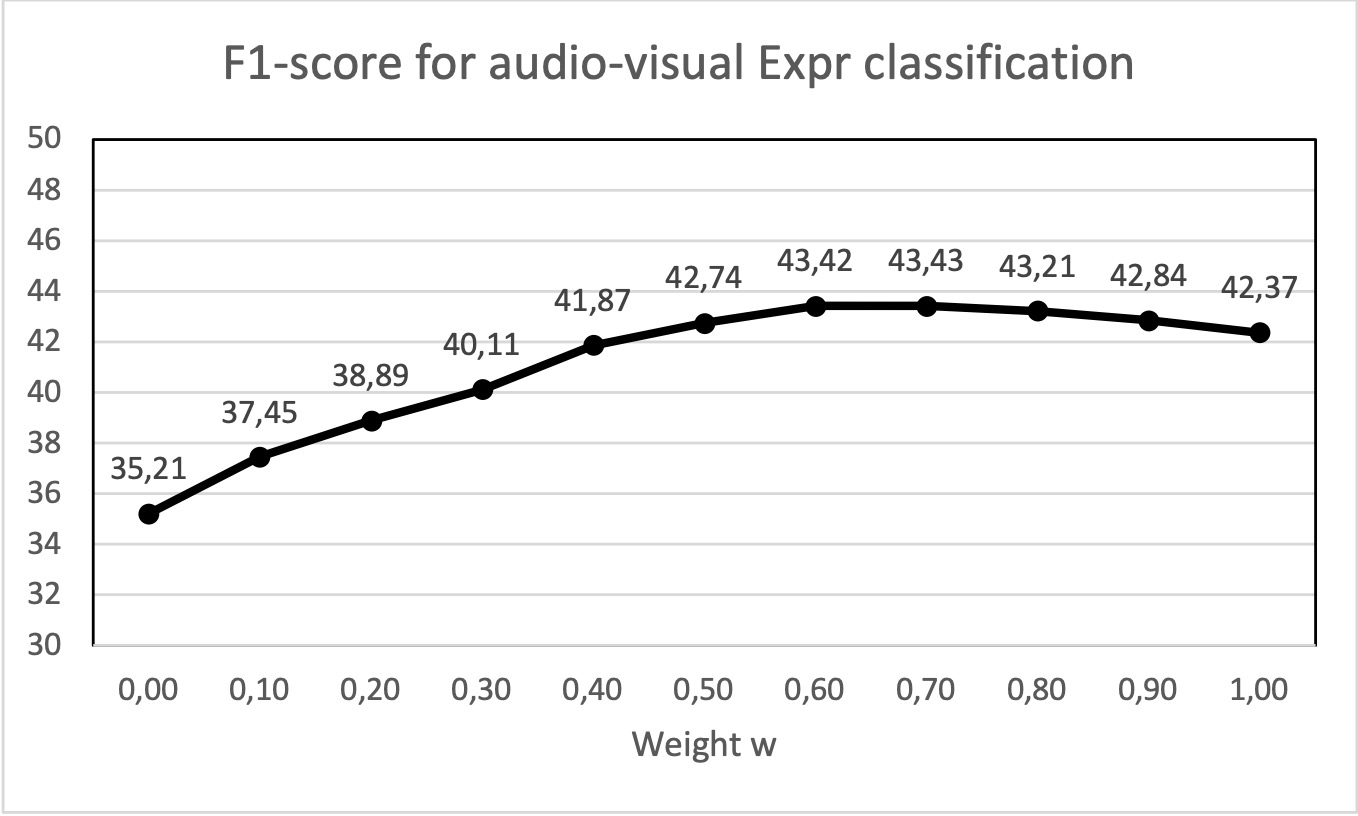}
 \caption{Dependence of F1-score for audio-visual Expr recognition on the weight $w$.}
 \label{fig:expr_weighting}
\end{figure}

\begin{figure}[t]
 \centering
 \includegraphics[width=0.95\linewidth]{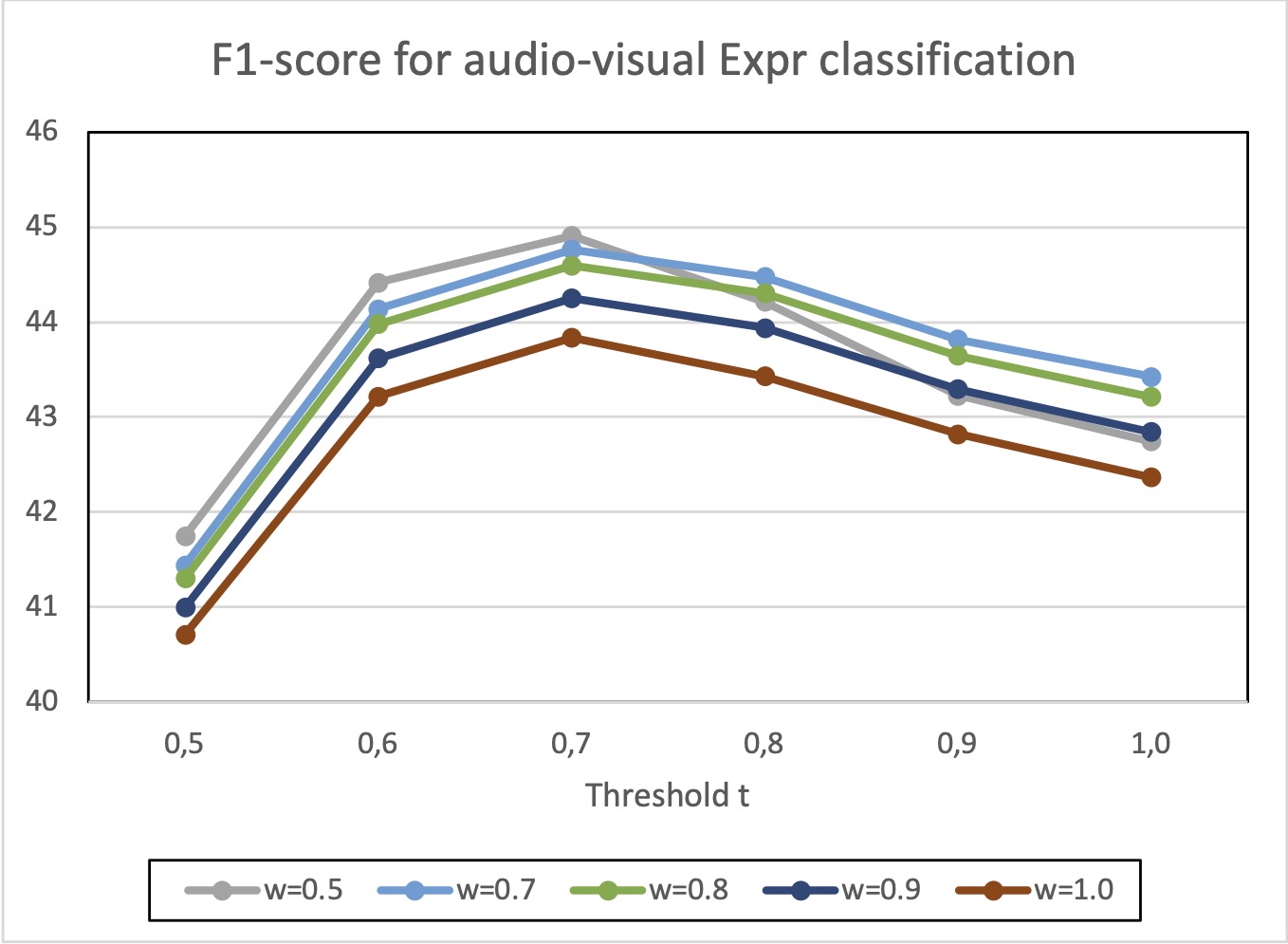}
 \caption{Dependence of F1-score for audio-visual Expr recognition on the filtering threshold $t$.}
 \label{fig:expr_audiovideo_thresholding}
\end{figure}

Fig.~\ref{fig:expr_smoothing} shows the dependence of the F1-score on the smoothing kernel size $k$, illustrating the impact of temporal smoothing on expression recognition. The dependence of the F1-score for video modality on the filtering threshold $t$ is presented in Fig.~\ref{fig:expr_video_thresholding}. Here, the top F1-score 43.84\% is obtained when $t=0.7$.

Fig.~\ref{fig:expr_weighting} further analyzes the effect of blending weight hyperparameter $w$ in the audiovisual recognition pipeline, demonstrating the optimal balance between modalities. Finally, as shown in Fig.~\ref{fig:expr_audiovideo_thresholding}, the filtering threshold $t$ significantly impacts the classification results, demonstrating the benefits of preprocessing techniques for improving recognition performance. These results confirm that frame filtering, temporal smoothing, and multimodal fusion significantly enhance the accuracy and robustness of FER.

\subsection{Emotional Mimicry Intensity Estimation}
\begin{table*}
\resizebox{\textwidth}{!}{
 \begin{tabular}{cccccccp{1cm}cc}
 \toprule
 Modality & Model & Features & PCC $\overline \rho$ & Admiration & Amusement & Determination & Empathic Pain & Excitement & Joy \\
 \midrule
Faces & \multicolumn{2}{c}{Baseline ViT~\cite{kollias20246th}} & 0.09 & -& -& -& -& -& -\\
Audio & \multicolumn{2}{c}{wav2Vec2~\cite{kollias20246th}}& 0.24& -& -& -& -& -& -\\
Audio+Video & \multicolumn{2}{c}{ViT+wav2Vec2~\cite{kollias20246th}} & 0.25& -& -& -& -& -& -\\
Audio+Video & \multicolumn{2}{c}{ResNet18+AUs+Wav2Vec2.0~\cite{yu2024efficient}} & 0.3288& -& -& -& -& -& -\\
Audio & \multicolumn{2}{c}{Wav2Vec2.0+VAD~\cite{Hallmen_2024_CVPR}} & 0.389& -& -& -& -& -& -\\
Audio+Video & \multicolumn{2}{c}{EmoViT+HuBERT+ChatGLM3~\cite{Qiu_2024_CVPR}} & 0.5851&  0.7155  & 0.6159 & 0.6303 & 0.3488 & 0.6174 & 0.5793\\
\hline
  & MT-& Embeddings (mean) & 0.1644 & 0.0379 & 0.2314 & 0.1387 & 0.0781 & 0.2334 & 0.2672 \\
Faces & EmotiMobile- & Embeddings (STAT) & 0.1683 & 0.0433 & 0.2459 & 0.1347 & 0.0779 & 0.2382 & 0.2699 \\
  & ViT& Scores (mean) & 0.1642 & 0.0321 & 0.2484 & 0.1490 & 0.0674 & 0.2399 & 0.2481 \\
  & & Scores (STAT) & 0.1776 & 0.0619 & 0.2623 & 0.1247 & 0.0773 & 0.2544 & 0.2849 \\ \hline
  & MT-& Embeddings (mean) & 0.1518 & 0.0215 & 0.2288 & 0.1140 & 0.0692 &  0.2299 & 0.2476 \\
Faces & Emoti- & Embeddings (STAT) & 0.1646 & 0.0557 & 0.2380 & 0.1303 & 0.0703 & 0.2325 & 0.2605 \\
 & MobileFaceNet & Scores (mean) & 0.1667 & 0.0276 & 0.2367 & 0.1336 & 0.0807 & 0.2516 & 0.2699 \\
  & & Scores (STAT) & 0.1732 & 0.0285 & 0.2498 & 0.1318 & 0.097 & 0.2543 & 0.2776 \\ \hline
Audio & wav2vec 2.0  & Embeddings (mean) & 0.1514 & 0.2153 & 0.11760 & 0.1834 & 0.1426 & 0.1275 & 0.1219 \\
 & & Embeddings (STAT) & 0.2311 & 0.3006 & 0.1659 & 0.2559 & 0.3198 &0.1844 & 0.1602 \\ \hline
Audio & HuBERT  & Embeddings (mean) & 0.3045 & 0.3644 & 0.3179 & 0.2987 & 0.3111 & 0.2848 & 0.2499 \\
 & & Embeddings (STAT) & 0.2729 & 0.3585 & 0.2454 & 0.2432 & 0.3065 & 0.2566 & 0.2275 \\ \hline
Text & RoBERTa  & Embeddings (mean) & 0.3763 & 0.4558 & 0.3328 &  0.3427 &  0.4964 &  0.3201 & 0.3099 \\
 & (emotional) & Embeddings (STAT) & 0.3697 & 0.4501 & 0.3326 &  0.3173 &  0.4843 &  0.3331 & 0.3008 \\ \hline
Text & SentenceBERT & Embeddings (mean) & 0.3756 & 0.4710 & 0.3585 & 0.3589 & 0.4347 & 0.3186 & 0.3119 \\
Text & Sentence all-MiniLM & Embeddings (mean) & 0.3537 & 0.4552 & 0.3076 & 0.3371 & 0.4111 & 0.3093 & 0.3023 \\ 
Text & SentenceE5 & Embeddings (mean) & 0.3777 & 0.4719 & 0.3497 & 0.3417 & 0.4553 & 0.3293 & 0.3184 \\ 
Text & OpenAI (small) & Embeddings (mean) & 0.4011 & 0.5113 & 0.3825 & 0.3687 & 0.4847 & 0.3363 & 0.3233 \\
Text & GigaChat & Embeddings (mean) & 0.4001 & 0.4888 & 0.3648 & 0.3732 & 0.4862 & 0.3726 & 0.3151 \\ \hline
Audio +& \multicolumn{2}{c}{wav2vec 2.0 + MT-EmotiMobileViT}& 0.2829 & 0.3011 & 0.2968 & 0.2595 & 0.3074 & 0.3171 & 0.2152 \\
Video & \multicolumn{2}{c}{wav2vec 2.0 + MT-EmotiMobileFaceNet}& 0.2898 & 0.3041 & 0.3004 & 0.2584 & 0.3148 & 0.3160 & 0.2452 \\
\hline
Text + Audio&\multicolumn{2}{c}{RoBERTa + HuBERT, blending} & 0.3974 & 0.4644 & 0.3727 & 0.3814 & 0.4755 & 0.3477 & 0.3430 \\
Text + Video&\multicolumn{2}{c}{RoBERTa + MT-EmotiMobileViT, blending} & 0.4028 & 0.4379 & 0.3911 & 0.3580 & 0.4802 & 0.3656 & 0.3840 \\
Text + Video&\multicolumn{2}{c}{RoBERTa + MT-EmotiMobileFaceNet, blending} & 0.4074 & 0.4408 & 0.3877 & 0.3632 & 0.5064 & 0.3650 & 0.3814 \\
Text + Video + Audio&\multicolumn{2}{c}{RoBERTa + MT-EmotiMobileViT + HuBERT, blending} & 0.4192 & 0.4603 & 0.4104 & 0.3844 & 0.4935 & 0.3802 & 0.3864 \\
Text + Video + Audio&\multicolumn{2}{c}{RoBERTa + MT-EmotiMobileFaceNet + HuBERT, blending} & 0.4223 & 0.4620 & 0.4095 & 0.3848 & 0.4936 & 0.3876 & 0.3965 \\
\hline
Text + Audio&\multicolumn{2}{c}{OpenAI + HuBERT, blending} & 0.4125& 0.5114& 0.4006& 0.3776& 0.4863& 0.3645& 0.3344 \\
Text + Video&\multicolumn{2}{c}{OpenAI + MT-EmotiMobileViT, blending} & 0.4103 & 0.4932 & 0.4090 & 0.3327 & 0.4571 & 0.3771 & 0.3930 \\
Text + Video&\multicolumn{2}{c}{OpenAI + MT-EmotiMobileFaceNet, blending} & 0.3887 & 0.4779 & 0.3593 & 0.3369 & 0.4373 & 0.3452 & 0.3758 \\
Text + Video + Audio&\multicolumn{2}{c}{OpenAI + MT-EmotiMobileViT + HuBERT, blending} & 0.4451 & 0.5203 & 0.4461 & 0.3938 & 0.5054 & 0.4015 & 0.4036 \\
Text + Video + Audio&\multicolumn{2}{c}{OpenAI + MT-EmotiMobileFaceNet + HuBERT, blending} & 0.4225 & 0.5073 & 0.4044 & 0.3696 & 0.4801 & 0.3822 & 0.3912 \\
\hline
Text + Audio&\multicolumn{2}{c}{GigaChat + HuBERT, early} & 0.4011& 0.4794& 0.3733& 0.4182& 0.4449& 0.3516& 0.3395 \\
Text + Audio&\multicolumn{2}{c}{GigaChat + HuBERT, blending} & 0.4106 & 0.4955 & 0.3912 & 0.3881 & 0.4619 & 0.3922 & 0.3348 \\
Text + Video&\multicolumn{2}{c}{GigaChat + MT-EmotiMobileViT, blending} & 0.4103 & 0.4932 & 0.4090 & 0.3327 & 0.4571 & 0.3771 & 0.3930 \\
Text + Video&\multicolumn{2}{c}{GigaChat + MT-EmotiMobileFaceNet, early} & 0.4199 & 0.4906 & 0.3988 & 0.4121 & 0.4420 & 0.3876 & 0.3883 \\
Text + Video&\multicolumn{2}{c}{GigaChat + MT-EmotiMobileFaceNet, blending} & 0.4231 & 0.4897 & 0.4107 & 0.3806 & 0.4824 & 0.4131 & 0.3618 \\
Text + Video + Audio&\multicolumn{2}{c}{GigaChat + MT-EmotiMobileViT + HuBERT, blending} & \bf 0.4460 & 0.5197 & 0.4491 & 0.3916 & 0.4981 & 0.4192 & 0.3983 \\ 
Text + Video + Audio&\multicolumn{2}{c}{GigaChat + MT-EmotiMobileFaceNet + HuBERT, early} & 0.4204 & 0.4823 &  0.4013 & 0.4267 & 0.4496 & 0.3765 & 0.3863 \\
Text + Video + Audio&\multicolumn{2}{c}{GigaChat + MT-EmotiMobileFaceNet + HuBERT, blending} & 0.4338 & 0.4955 & 0.4332 & 0.3923 & 0.4640 & 0.4282 & 0.3898 \\ 
\bottomrule
 \end{tabular}
 }
 \caption{Pearson's correlation for EMI Estimation on the Hume-Vidmimic2's validation set.}
 \label{tab:emi}
\end{table*}

For the EMI Estimation task, we evaluated different feature extraction and fusion methods on the Hume-Vidmimic2 validation set. Table~\ref{tab:emi} presents Pearson’s correlation coefficients (PCC) for various models across different emotions.

The baseline ViT model for facial features alone achieved a correlation of 0.09, while wav2vec 2.0 audio embeddings~\cite{baevski2020wav2vec} significantly improved performance to 0.24. Combining visual and audio embeddings (ViT + wav2vec 2.0) increased the correlation to 0.25, demonstrating the importance of multimodal integration.

From our results, one can notice that the impact of visual modality is very small. However, our best model, GigaChat text embeddings blended with HuBERT audio~\cite{hsu2021hubert} and MT-EmotiMobileViT facial video~\cite{Savchenko_2024a_CVPR} features, achieved a correlation of 0.4460, marking a significant improvement over unimodal approaches. 

\subsection{Ambivalence/Hesitancy Recognition}

\begin{table}
 \centering
 \resizebox{\columnwidth}{!}{
 \begin{tabular}{cccc}
 \toprule
 Method & Modality & F1-score, \% \\
 \midrule
MT-EmotiMobileFaceNet, Scores & Faces & 65.35 \\
MT-EmotiMobileFaceNet, Features & Faces & 64.57 \\
\hline
wav2vec 2.0 & Audio  & 68.50\\ 
HuBERT & Audio  & 69.29\\ 
\hline
OpenAI small & Text & 70.08\\
Gigachat & Text & 73.23 \\
RoBERTa (emotional) & Text & 77.16 \\
 \bottomrule
 \end{tabular}
 }
 \caption{Video-level Ambivalence/Hesitancy Recognition Accuracy on the BAH’s validation set.}
 \label{tab:bah_video}
\end{table}

\begin{table*}
 \centering
 \begin{tabular}{cccc}
 \toprule
 Method & Modality & F1-score, \% \\
 \midrule
 Baseline~\cite{Kollias2025} & Faces + Audio + Text & 70.0 \\
\hline
DDAMFN, Scores & Faces & 67.97 \\
DDAMFN, Features & Faces & 68.36 \\
EmotiEffNet-B0, Scores& Faces & 69.75 \\ 
EmotiEffNet-B0, Features& Faces & 67.66 \\ 
MT-EmotiMobileViT, Scores & Faces & 69.81 \\
MT-EmotiMobileViT, Features & Faces & 67.69 \\
MT-DDAMFN, Scores & Faces & 68.76 \\
MT-DDAMFN, Features & Faces & 68.02 \\
MT-EmotiEffNet, Scores & Faces & 68.32 \\
MT-EmotiEffNet, Features & Faces & 68.46 \\
MT-EmotiMobileFaceNet, Scores & Faces & 70.59 \\
MT-EmotiMobileFaceNet, Features & Faces & 68.36 \\
\hline
wav2vec 2.0 & Audio  & 67.66\\ 
HuBERT & Audio  & 68.57\\ 
RoBERTa (emotional) & Text & 70.70 \\
\hline
MT-EmotiMobileFaceNet, smoothing & Faces  & 72.01\\
MT-EmotiMobileFaceNet, smoothing+ filtering & Faces  & 72.11\\
HuBERT+MT-EmotiMobileFaceNet, blending& Audio + Faces & 71.51 \\
RoBERTa+MT-EmotiMobileFaceNet, blending& Text + Faces & 71.45 \\
RoBERTa + HuBERT + MT-EmotiMobileFaceNet, blending& Text + Audio + Faces & 72.12 \\
RoBERTa + HuBERT + MT-EmotiMobileFaceNet, fusion& Text + Audio + Faces & 72.01 \\ 
RoBERTa+MT-EmotiMobileFaceNet, fusion& Text + Faces & 72.26 \\
RoBERTa+MT-EmotiMobileFaceNet, smoothing & Text + Faces  & 73.31\\
RoBERTa+MT-EmotiMobileFaceNet, smoothing + filtering & Text + Faces  & 73.73\\
 \bottomrule
 \end{tabular}
 \caption{Frame-level Ambivalence/Hesitancy Recognition Challenge Results on the BAH’s validation set.}
 \label{tab:bah_frame}
\end{table*}

We evaluated different multimodal strategies on the BAH validation set for the AH Recognition task. At first, we performed intermediate experiments with the global prediction of AH for the entire video. Table~\ref{tab:bah_video} contains the F1-score for this task. Here, the top performance is achieved by text modality using RoBERTa emotional embeddings, so we used it in the main experiment if filtering is applied

Table~\ref{tab:bah_frame} presents the main results, F1-scores achieved using various feature extraction and fusion methods. Baseline approaches using video (MT-EmotiMobileFaceNet), audio (HuBERT), and text embeddings (RoBERTa) achieved scores of 70.59\%, 68.57\%, and 70.70\%, respectively. With smoothing and filtering, the best unimodal visual model, MT-EmotiMobileFaceNet~\cite{Savchenko_2024a_CVPR}, achieved an F1-score of 72.11\%.

Our best-performing multimodal model, which combines RoBERTa text embeddings~\cite{liu2019roberta} with MT-EmotiMobileFaceNet~\cite{Savchenko_2024a_CVPR} and smoothing, achieved an F1-score of 73.73\%, confirming the effectiveness of integrating textual and visual modalities. 

\section{Conclusion}
In this paper, we introduced the multimodal approach (Fig.~\ref{fig:pipeline}) that leverages facial, audio, and textual modalities in a computationally efficient manner. We use pre-trained modes to extract features without the need for their fine-tuning on each task. As a result, we significantly improved the performance of three considered tass (EXPR classification, EMI estimation, and AH recognition) compared to existing baselines~\cite{Kollias2025}. In addition, we used our best models from the ABAW-6 competition in valence-arousal estimation and action unit detection tasks. The source code to reproduce the experiments is publicly available\footnote{\url{https://github.com/av-savchenko/EmotiEffLib/tree/main/training_and_examples/ABAW/ABAW8}}. Our most interesting innovation included filtering of reliable facial expressions obtained by a pre-trained model without refinement on given training sets (Table~\ref{tab:expr}) and usage of global video classifiers and interpolation of acoustic and text embeddings to improve the quality of AH prediction (Table~\ref{tab:bah_frame}). 

Our work contributes to the ongoing advancement of affective computing research by addressing challenges related to multimodal fusion, domain-specific generalization, and real-time inference. The findings from this study highlight the importance of leveraging multimodal signals and intelligent data selection strategies to achieve state-of-the-art performance in affective behavior analysis. The ability to analyze affective behavior efficiently has significant implications for real-world applications, including emotion-aware virtual agents, adaptive learning systems, and automated mental health assessment tools.

\textbf{Acknowledgements}. The article was prepared within the framework of the Basic Research Program at the National Research University Higher School of Economics (HSE).

{
    \small
    \bibliographystyle{ieeenat_fullname}
    \bibliography{main}
}


\end{document}